# R2H-Diff: Guided Spectral Diffusion Model for RGB-to-Hyperspectral Reconstruction

Songyu Ding, Rongqiang Zhao*, Mingchun Sun, Jie Liu *Fellow, IEEE*

***Abstract*—RGB-to-hyperspectral image reconstruction is a highly ill-posed inverse problem, since multiple plausible spectral distributions may correspond to the same RGB observation. Existing regression-based methods usually learn a deterministic mapping, which limits their ability to model reconstruction uncertainty and often leads to over-smoothed spectral responses. Although diffusion models provide strong distribution modeling capability, their direct application to hyperspectral reconstruction remains challenging due to the high spectral dimensionality, strong inter-band correlations, and strict requirement for spectral fidelity. To this end, we propose R2H-Diff, an efficient diffusion-based framework tailored for RGB-to-HSI reconstruction. Specifically, R2H-Diff formulates spectral recovery as a conditional iterative refinement process, enabling progressive reconstruction under RGB guidance. We proposed a Guided Spectral Refinement Module for RGB-conditioned feature fusion and a Hyperspectral-Adaptive Transposed Attention module for efficient spatial–spectral dependency modeling. Furthermore, a normalization-free denoising backbone is adopted to preserve spectral amplitude consistency, while a task-adapted linear noise schedule enables high-quality reconstruction with only five denoising steps. Extensive experiments on NTIRE2022, CAVE, and Harvard demonstrate that R2H-Diff achieves a favorable balance between reconstruction quality and computational efficiency. Notably, on NTIRE2022, R2H-Diff obtains 35.37 dB PSNR with a sub-million-parameter model of 0.58M parameters and 12.25G FLOPs, achieving the lowest model complexity among the evaluated methods while maintaining strong reconstruction fidelity.**

## I. Introduction

HYPERSPECTRAL images (HSIs) provide dense and contiguous spectral responses beyond conventional RGB imaging, enabling accurate material representation and fine-grained scene understanding. Benefiting from such rich spectral information, HSIs have been widely used in remote sensing, precision agriculture, environmental monitoring, semantic analysis, target detection, and multimedia perception tasks [1], [2], [3], [4], [5], [6]. However, acquiring high-quality HSIs usually requires specialized spectral sensors and scanning mechanisms, which are expensive, time-consuming, and difficult to deploy in dynamic or real-time imaging scenarios. In contrast, RGB images can be readily captured by low-cost commercial cameras. Therefore, RGB-to-HSI reconstruction has attracted increasing attention as a practical and cost-effective approach for recovering high-fidelity spectral information from RGB observations [7], [8], [9], [10]. This task provides an important bridge between economical RGB imaging and information-rich hyperspectral sensing, while also posing a challenging ill-posed inverse problem due to severe spectral information loss during RGB image formation.

*Corresponding author: Rongqiang Zhao@zhaorq@hit.edu.cn.
The authors are with the Faculty of Computing, Harbin Institute of Technology, Harbin 150001, China.
The authors are with the State Key Laboratory of Smart Farm Technologies and Systems, Harbin 150001, China.

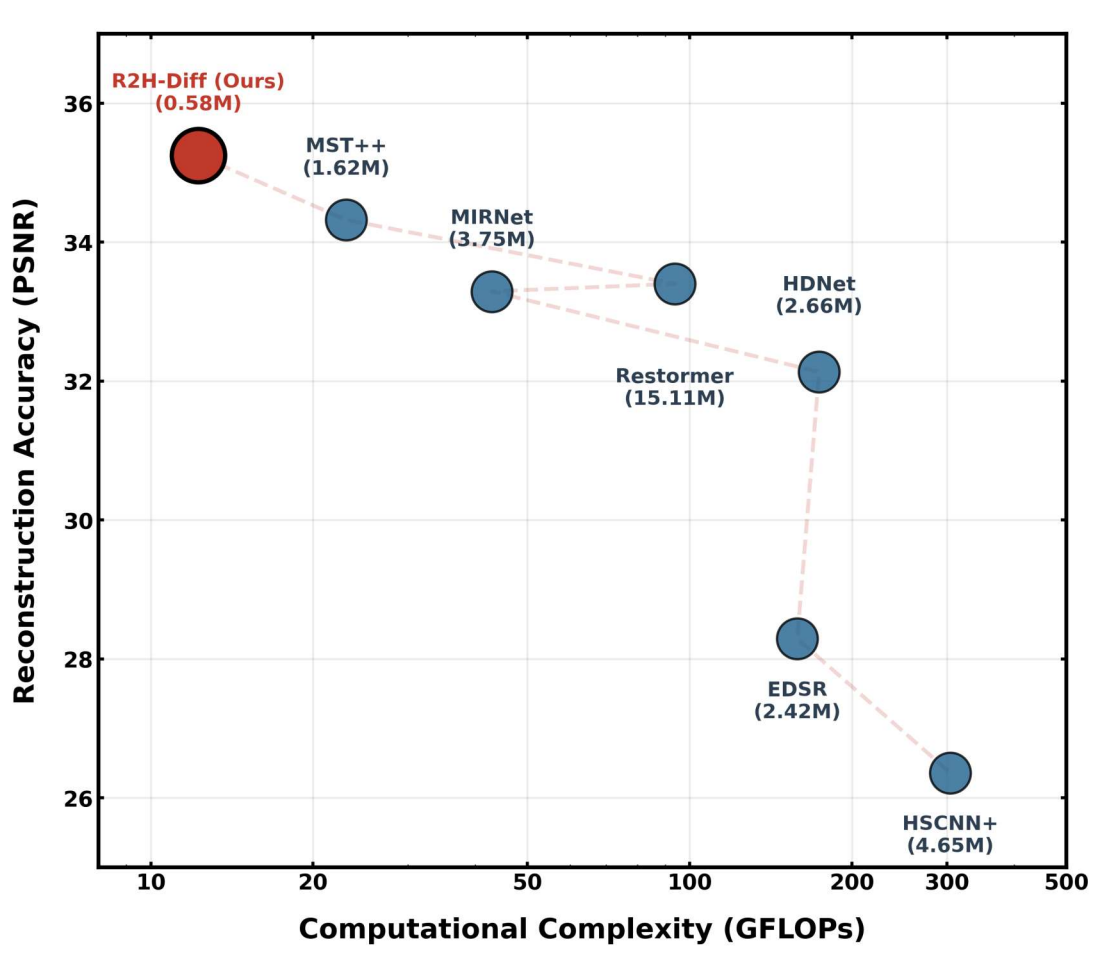

Fig. 1: Complexity–performance comparison on NTIRE2022. PSNR, GFLOPs, and parameter counts are reported for state-of-the-art spectral reconstruction methods. For R2H-Diff, the reported FLOPs are computed over the complete five-step denoising inference process.

Despite its practical value, RGB-to-HSI reconstruction is highly ill-posed, since the broad and overlapping camera response functions cause severe spectral information loss and make multiple spectral distributions correspond to the same RGB observation. Traditional methods typically regularize this inverse problem with handcrafted priors, such as sparsity, low-rankness, tensor structures, and frequency-domain constraints [11], [12], [13]. Although interpretable, these methods often rely on complex optimization and have limited representation capability for diverse spectral distributions [14], [15]. With the development of deep learning, CNN-based methods learn an end-to-end RGB-to-HSI mapping and effectively capture local spatial–spectral correlations [16], [17], [18]. Image restoration architectures such as MIRNet and Restormer further enhance feature representation through multi-scale aggregation and attention mechanisms [19], [20]. More recently, Transformer-based methods have been introduced to model long-range dependencies and global spectral interactions, showing promising performance in spectral recovery and compressive spectral imaging [21], [22], [23], [24], [25].

Nevertheless, most existing CNN- and Transformer-based methods follow a deterministic regression paradigm. Under pixel-wise losses, they tend to estimate an averaged spectral solution, which may suppress subtle spectral fluctuations and produce over-smoothed spectral responses. Generative models provide a promising alternative by learning the underlying data distribution rather than a single deterministic mapping. Early generative models, including GANs and VAEs, have demonstrated the ability to model complex data manifolds [26], [27]. Diffusion models further improve generation and restoration quality through iterative denoising, and have shown strong potential in image super-resolution, editing, restoration, and high-resolution synthesis [28], [29], [30], [31], [32], [33], [34]. In hyperspectral and computational imaging, recent diffusion-based methods have also been explored for spectral super-resolution, restoration, and reconstruction [35], [36], [37]. However, generic diffusion models remain difficult to apply to RGB-to-HSI reconstruction due to high spectral dimensionality, inter-band correlations, amplitude-sensitive spectra, and costly sampling.

To address these challenges, we propose R2H-Diff, an efficient diffusion-based framework tailored for RGB-to-HSI reconstruction. R2H-Diff formulates spectral recovery as a conditional iterative refinement process, progressively reconstructing high-fidelity HSIs under RGB guidance. To adapt diffusion modeling to hyperspectral data, we design a hyperspectral-adaptive transposed attention module (HATA) for efficient spatial–spectral dependency modeling and a Guided Spectral Refinement Module (GSRM) for RGB-conditioned feature fusion. In addition, a normalization-free denoising backbone is adopted to avoid spectral amplitude distortion, and a gradient consistency constraint is introduced to preserve fine structural details. With a task-adapted linear noise schedule, R2H-Diff achieves high-quality reconstruction with only five denoising steps, providing an effective balance between reconstruction accuracy and inference efficiency. The main contributions of this paper are summarized as follows:

1) We proposed R2H-Diff, an efficient diffusion-based framework that reformulates RGB-to-HSI reconstruction as a conditional iterative refinement process, achieving high-quality reconstruction with only five denoising steps.
2) We proposed a Hyperspectral-Adaptive Transposed Attention module to capture spatial–spectral dependencies with low computational complexity, enabling effective modeling of high-dimensional hyperspectral features.
3) We proposed a Guided Spectral Refinement Module to enhance RGB-conditioned spectral feature fusion. In addition, a normalization-free backbone and gradient consistency constraint are adopted to preserve spectral fidelity and structural details.

## II. Related Work

### A. Hyperspectral Image Reconstruction

Hyperspectral image reconstruction aims to recover high-fidelity spectral information from limited or degraded observations, such as RGB images, compressed measurements, and low-resolution inputs. Early methods mainly rely on hand-crafted priors, including sparsity, low-rank modeling, tensor decomposition, frequency-domain constraints, and physical imaging priors, to regularize this ill-posed inverse problem [9], [6], [11], [13], [12]. Although interpretable, these methods are usually limited by manually designed assumptions and iterative optimization costs. Deep learning-based methods have therefore become dominant by learning end-to-end mappings from degraded observations to hyperspectral outputs [16], [17], [18], [38]. Subsequent studies further improve reconstruction quality through dual-domain modeling, spectral unmixing priors, self-supervised restoration, and frequency-domain learning [39], [40], [41], [42]. Recently, Transformer-based architectures have been introduced to capture long-range spatial–spectral dependencies and cross-band correlations, achieving promising performance in spectral recovery and compressive hyperspectral imaging [10], [43], [23], [22], [25]. However, most CNN- and Transformer-based methods still follow a deterministic regression paradigm, which may produce averaged spectral estimates and suppress subtle spectral variations. This motivates reconstruction models with stronger distribution modeling capability and improved spectral fidelity.

### B. Generative Models for Image Reconstruction

Generative models provide effective data priors for ill-posed image reconstruction by modeling the underlying data distribution rather than a single deterministic mapping. Early generative models, such as GANs and VAEs, have shown the ability to learn complex data manifolds [26], [27]. More recently, diffusion models have attracted increasing attention due to their stable training, strong generation quality, and flexible conditional modeling ability [28], [29], [30], [31]. By progressively denoising corrupted inputs, diffusion models have been successfully applied to image super-resolution, restoration, enhancement, and detail refinement [32], [44], [45]. Beyond image restoration, generative augmentation has also been investigated for industrial data generation, while diffusion-based generation has been extended to time-series synthesis [46]. In hyperspectral and computational imaging, diffusion-based priors have been explored for spectral super-resolution, hyperspectral restoration, and hyperspectral reconstruction [47], [35], [36], [37]. Nevertheless, directly applying generic diffusion models to RGB-to-HSI reconstruction remains challenging because hyperspectral data have high spectral dimensionality, strong inter-band correlations, amplitude-sensitive responses, and high sampling costs. Therefore, an efficient diffusion framework tailored to RGB-to-HSI reconstruction is needed to balance spectral fidelity, conditional guidance, and inference efficiency.

## III. Proposed Method

R2H-Diff formulates RGB-to-HSI reconstruction as an RGB-guided conditional diffusion process. As shown in Fig. 2, GSRM provides conditional spectral guidance, while the HATA-based U-Net progressively refines noisy spectral states to recover the clean HSI. The model is trained with reconstruction and gradient consistency losses.

### A. Denoising Diffusion

R2H-Diff formulates RGB-to-HSI reconstruction as an RGB-guided conditional diffusion process. Following

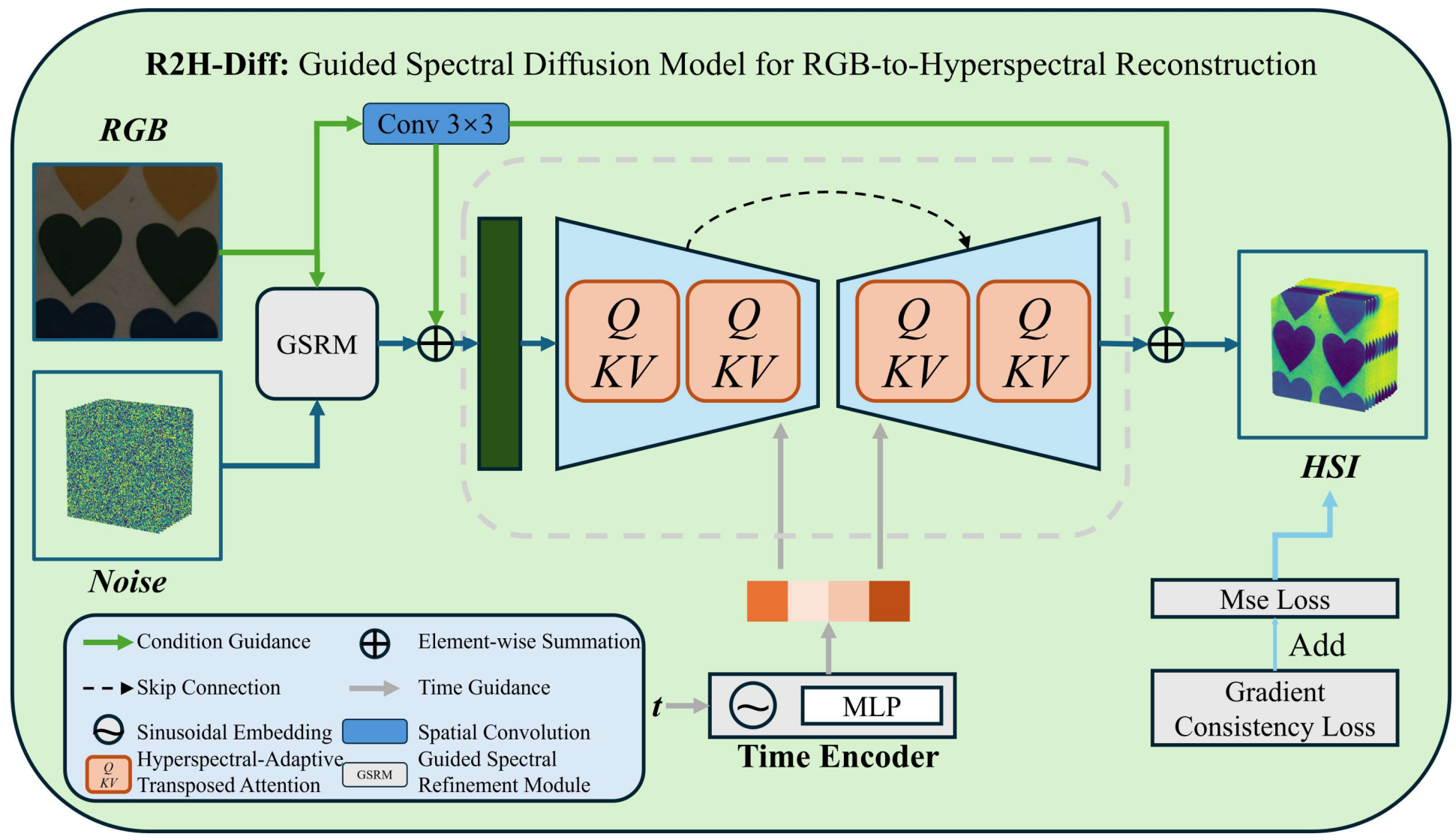

Fig. 2: Overall framework of R2H-Diff. GSRM provides RGB-conditioned guidance, and the HATA-based denoising U-Net progressively reconstructs the HSI under timestep guidance.

DDPM [48], the forward process gradually perturbs the ground-truth HSI $\mathbf{x}_0$ into noisy spectral states $\mathbf{x}_t$:

$$q(\mathbf{x}_t \mid \mathbf{x}_0) = \mathcal{N}\left(\mathbf{x}_t; \sqrt{\bar{\alpha}_t}\mathbf{x}_0, (1-\bar{\alpha}_t)\mathbf{I}\right), \tag{1}$$

where $\bar{\alpha}_t = \prod_{i=1}^{t} \alpha_i$ and $\alpha_t = 1 - \beta_t$. Equivalently, $\mathbf{x}_t$ can be sampled as

$$\mathbf{x}_t = \sqrt{\bar{\alpha}_t}\mathbf{x}_0 + \sqrt{1-\bar{\alpha}_t}\boldsymbol{\epsilon}, \quad \boldsymbol{\epsilon} \sim \mathcal{N}(\mathbf{0}, \mathbf{I}). \tag{2}$$

The reverse process is conditioned on the RGB observation $\tilde{\mathbf{x}}$:

$$p_\theta(\mathbf{x}_{t-1} \mid \mathbf{x}_t, \tilde{\mathbf{x}}). \tag{3}$$

Instead of predicting the noise term, we directly predict the clean hyperspectral image:

$$\hat{\mathbf{x}}_\theta = \hat{\mathbf{x}}_\theta(\mathbf{x}_t, \tilde{\mathbf{x}}, t). \tag{4}$$

This $\mathbf{x}_0$-prediction strategy is more consistent with the reconstruction objective and better exploits the spatial correspondence between the RGB guidance and the target HSI.

The basic diffusion reconstruction objective is defined as

$$\mathcal{L}_{\text{diff}} = \mathbb{E}_{\mathbf{x}_0,\tilde{\mathbf{x}},\boldsymbol{\epsilon},t} \left\| \mathbf{x}_0 - \hat{\mathbf{x}}_\theta(\mathbf{x}_t, \tilde{\mathbf{x}}, t) \right\|_2^2. \tag{5}$$

The final training objective with gradient consistency regularization is introduced in Sec. III-B4.

During inference, we adopt deterministic DDIM sampling by setting $\sigma_t = 0$ [49]. The transition from $\mathbf{x}_t$ to $\mathbf{x}_{t-1}$ is given by

$$\mathbf{x}_{t-1} = \sqrt{\bar{\alpha}_{t-1}}\hat{\mathbf{x}}_\theta + \sqrt{1-\bar{\alpha}_{t-1}} \frac{\mathbf{x}_t - \sqrt{\bar{\alpha}_t}\hat{\mathbf{x}}_\theta}{\sqrt{1-\bar{\alpha}_t}}. \tag{6}$$

To support efficient few-step reconstruction, we use a task-adapted linear schedule on $\sqrt{\bar{\alpha}_t}$:

$$\sqrt{\bar{\alpha}_t} = 1 - (1-\delta)\frac{t}{T}, \quad t = 0, \ldots, T, \tag{7}$$

where $\delta$ is a small positive constant for numerical stability. The corresponding parameters are obtained by

$$\alpha_t = \frac{\bar{\alpha}_t}{\bar{\alpha}_{t-1}}, \quad \beta_t = 1 - \alpha_t. \tag{8}$$

This formulation enables deterministic few-step spectral refinement under RGB guidance.

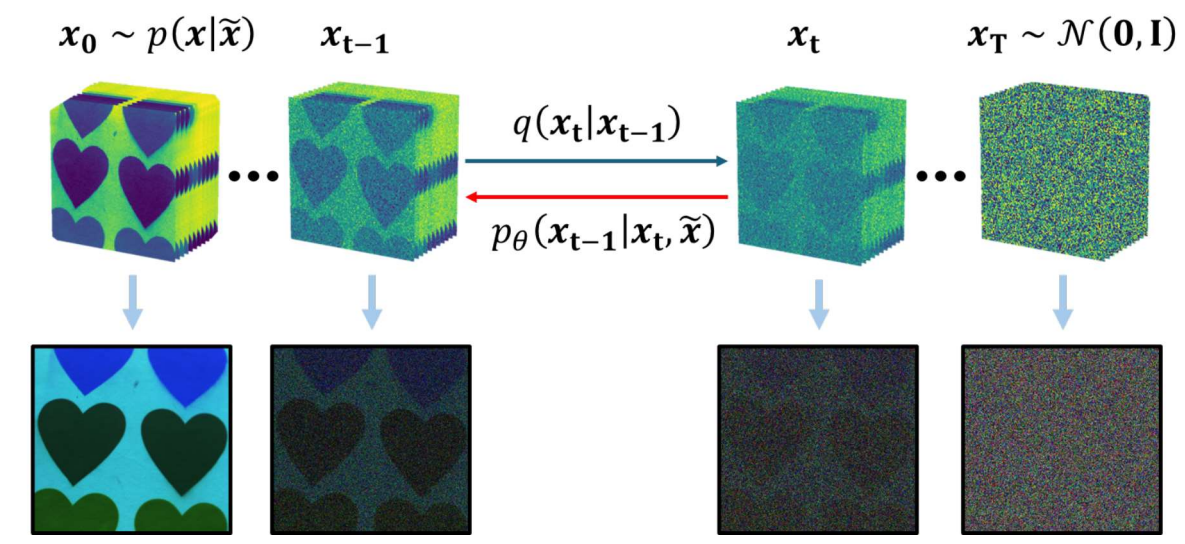

Fig. 3: Illustration of the diffusion framework. The spectral cube $\mathbf{x}_0$ is gradually perturbed into Gaussian noise $\mathbf{x}_T$ and then progressively reconstructed under RGB guidance $\tilde{\mathbf{x}}$.

### B. *Network Architecture*

Fig. 2 illustrates the architecture of R2H-Diff, which formulates RGB-to-HSI reconstruction as an iterative conditional diffusion process. The framework employs GSRM to extract spatial-spectral priors from RGB inputs, effectively bridging the dimensionality mismatch for conditional guidance. The core U-Net backbone utilizes Transposed Attention blocks to capture long-range dependencies while maintaining computational efficiency, with a Time Encoder providing step-wise guidance via sinusoidal embeddings. To facilitate convergence, a global residual connection provides a base spectral estimate, and the model is optimized using a joint loss of Mean Squared Error and Gradient Consistency to ensure both spectral accuracy and structural preservation.

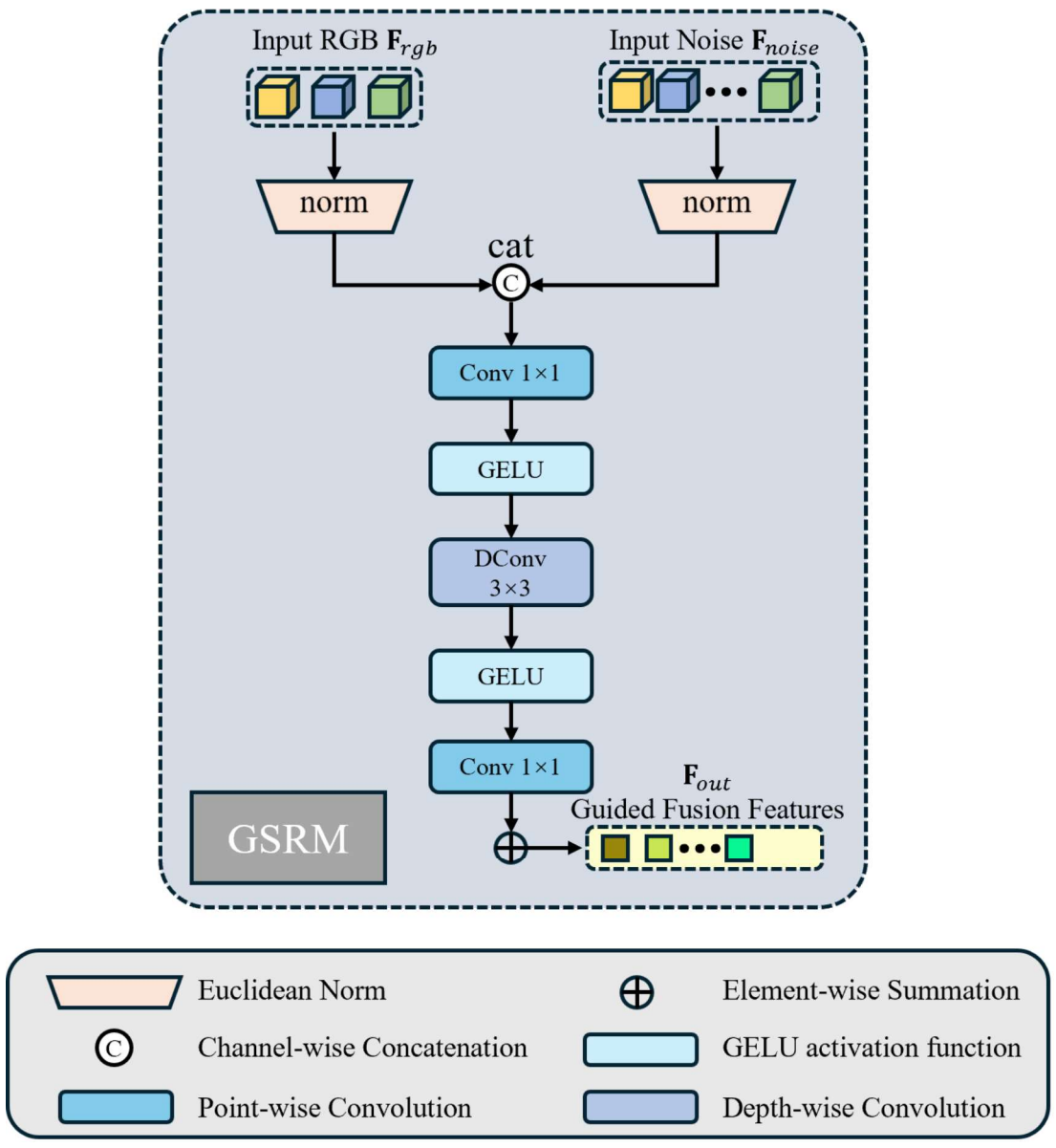

Fig. 4: Architecture of GSRM. It fuses RGB-conditioned priors and noise features using dual-branch normalization and lightweight convolutional encoding.

*1) Guided Spectral Refinement Module:* As illustrated in Fig. 4, GSRM adopts a dual-branch architecture to process the RGB condition $\mathbf{F}_{rgb}$ and the noise $\mathbf{F}_{noise}$. To enhance numerical stability, both inputs are regularized via Euclidean Normalization:

$$\hat{\mathbf{F}} = \frac{\mathbf{F}}{\|\mathbf{F}\|_2 + \epsilon}, \tag{9}$$

where $\epsilon$ is a small constant. The normalized features are then concatenated to form a joint representation $\mathbf{F}_0 = [\hat{\mathbf{F}}_{rgb}, \hat{\mathbf{F}}_{noise}]$.

The core of GSRM is a bottleneck structure designed for efficient feature encoding and modality fusion. Specifically, the joint representation $\mathbf{F}_0$ undergoes a $1 \times 1$ convolution for dimensionality transformation, followed by a GELU activation $\sigma(\cdot)$. To capture spatial context, a $3\times3$ depth-wise convolution (DWConv) is integrated. The final guided fusion features $\mathbf{F}_{out}$ are generated through a successive $1 \times 1$ convolutional layer:

$$\mathbf{F}_{out} = \mathrm{Conv}_{1\times1}(\sigma(\mathrm{DWConv}_{3\times3}(\sigma(\mathrm{Conv}_{1\times1}(\mathbf{F}_0))))), \tag{10}$$

which provides the integrated conditional guidance for the subsequent denoising backbone.

*2) Denoising Network:* The denoising network adopts a U-Net-based architecture specifically optimized for hyperspectral reconstruction. To strike a balance between performance and computational efficiency, we employ a lightweight configuration with a base channel dimension of $C = 31$ and channel multipliers of $(1, 1, 1)$, spanning three resolution scales. Each scale consists of two Residual Blocks (ResBlocks) followed by a spectral attention module. Long-range skip connections are incorporated between the encoder and decoder to preserve multi-scale spatial details.

To mitigate the difficulty of direct high-dimensional mapping, we formulate the reconstruction as a residual learning task. A coarse spectral estimate $\mathbf{x}_{\text{base}} \in \mathbb{R}^{H\times W\times B}$ is first derived from the RGB input $\mathbf{x}_{\text{rgb}}$ via a simple mapping function $\phi(\cdot)$, implemented as a $3 \times 3$ convolution:

$$\mathbf{x}_{\text{base}} = \phi(\mathbf{x}_{\text{rgb}}). \tag{11}$$

The denoising network $\mathcal{F}_\theta$ then predicts a spectral residual to refine this coarse estimate. The final reconstructed HSI $\hat{\mathbf{x}}_0$ is defined as:

$$\hat{\mathbf{x}}_0 = \mathbf{x}_{\text{base}} + \mathcal{F}_\theta(\mathbf{x}_t, \mathbf{x}_{\text{rgb}}, t), \tag{12}$$

where $t$ denotes the diffusion timestep. This residual formulation eases the high-dimensional reconstruction task and improves training stability.

The temporal information is integrated by encoding $t$ into a sinusoidal embedding $\mathbf{e}(t)$, which is subsequently transformed by a multi-layer perceptron (MLP) $\psi(\cdot)$ and injected into each ResBlock:

$$\mathbf{h}' = \mathbf{h} + \psi(\mathbf{e}(t)), \tag{13}$$

where $\mathbf{h}$ represents the intermediate feature map. Notably, we deviate from standard U-Net designs by removing normalization layers (e.g., GroupNorm) within the ResBlocks, retaining only $3 \times 3$ convolutions and SiLU activations. This design choice is predicated on the observation that spectral bands often exhibit independent radiometric characteristics; avoiding cross-channel normalization prevents spectral distortion and maintains the fidelity of narrow-band information. A single normalization layer is only retained in the final refinement stage to ensure numerical stability.

*3) Hyperspectral-Adaptive Transposed Attention:* In R2H-Diff, HATA is adapted for hyperspectral reconstruction by modeling channel-wise spectral dependencies and complementing them with local spatial cues from the LPE-based gated branch. To efficiently model global spectral dependencies, we adopt a Transposed Attention mechanism as illustrated in Fig. 5. Given an input $\mathbf{X} \in \mathbb{R}^{B\times C\times H\times W}$, a $1 \times 1$ pointwise convolution is first applied to project the channels from $C$ to $3 \times C$, followed by a *chunk* operation to generate the Query ($\mathbf{Q}$), Key ($\mathbf{K}$), and Value ($\mathbf{V}$). Unlike conventional attention, $\mathbf{Q}$ and $\mathbf{K}$ are reshaped to $\mathbb{R}^{B\times(HW)\times C}$ and L2-normalized to compute a channel-wise attention map of size $\mathbb{R}^{C\times C}$:

$$\mathrm{Attn}(\mathbf{Q}, \mathbf{K}) = \mathrm{Softmax}\left(\frac{\hat{\mathbf{Q}}^\top \hat{\mathbf{K}}}{\alpha}\right), \tag{14}$$

where $\hat{\mathbf{Q}}$ and $\hat{\mathbf{K}}$ denote the normalized features and $\alpha$ is a learnable scaling factor. This produces a channel-wise attention map of size $C \times C$, avoiding the quadratic spatial complexity of conventional self-attention. The computational cost is reduced to $\mathcal{O}(HWC^2)$, which is more efficient than spatial attention with $\mathcal{O}((HW)^2C)$ complexity.

Simultaneously, a Local Position Embedding (LPE) branch originates from $\mathbf{V}$, consisting of two $3 \times 3$ depth-wise convolutions and a GELU activation. This branch captures local spatial context and generates a gating mask through a Sigmoid function $\sigma$. The output is obtained via a gated fusion strategy:

$$\mathbf{X}_{out} = (\mathbf{X}_{attn} \otimes \sigma(\mathbf{X}_{lpe})) + \mathbf{X}_{lpe}, \tag{15}$$

where $\otimes$ denotes the Hadamard product and $\mathbf{X}_{attn}$ is the output of the attention mechanism. This design integrates global spectral correlations and local spatial cues without requiring absolute positional embeddings.

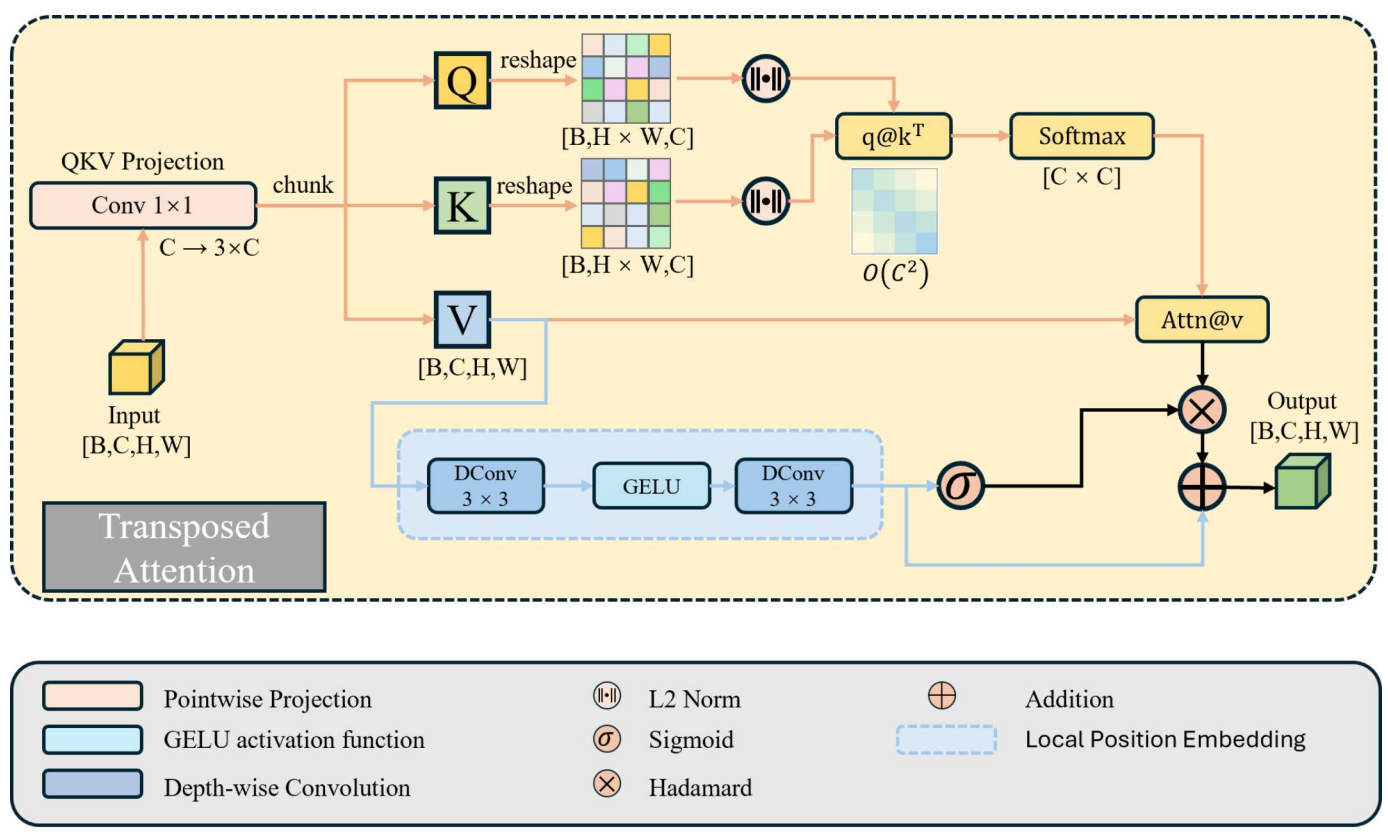

Fig. 5: Architecture of HATA. It models spectral dependencies through channel-wise self-attention and refines local spatial details using an LPE-based gated branch.

*4) Loss Function:* In alignment with our reformulated diffusion objective, the model is trained to directly predict the clean spectral data $\mathbf{x}_0$. The training objective is derived as the Mean Squared Error (MSE) between the ground truth and the network's estimate:

$$\mathcal{L}_{mse} = \mathbb{E}_{\mathbf{x}_0,\mathbf{x}_t,t} \left\| \mathbf{x}_0 - \hat{\mathbf{x}}_\theta(\mathbf{x}_t, t, \tilde{\mathbf{x}}) \right\|_2^2 . \tag{16}$$

Unlike conventional noise-prediction ($\epsilon$-prediction), this $\mathbf{x}_0$-regression target is more suitable for spectral reconstruction, as it leverages the strong spatial correlations between the RGB condition $\tilde{\mathbf{x}}$ and the target HSI $\mathbf{x}_0$ to accelerate convergence.

To further refine the high-frequency structural components, we augment the training with a **gradient consistency loss** $\mathcal{L}_{grad}$. This term constrains the structural discrepancy in the spatial gradient domain:

$$\mathcal{L}_{grad} = \sum_{d\in\{h,v\}} \|\nabla_d(\hat{\mathbf{x}}_0) - \nabla_d(\mathbf{x}_0)\|_2^2, \tag{17}$$

where $\nabla_h$ and $\nabla_v$ represent the horizontal and vertical Sobel operators, respectively. By penalizing gradient mismatches, this term effectively prevents over-smoothing and preserves sharp edge details. The overall objective is formulated as:

$$\mathcal{L}_{\text{total}} = \mathcal{L}_{\text{mse}} + \lambda \mathcal{L}_{\text{grad}}. \tag{18}$$

where we empirically set the balance hyperparameter $\lambda = 1$ to ensure a harmonious integration of pixel-level radiometric accuracy and structural fidelity. The detailed implementation of the proposed framework is summarized in Algorithms 1 and 2. Specifically, Algorithm 1 delineates the diffusion training process supervised by $\mathcal{L}_{total}$, while Algorithm 2 describes the iterative reverse sampling procedure for deterministic HSI reconstruction.

## IV. Experimental Results

### A. Experimental Settings

*1) Benchmark Datasets:* To evaluate the performance of our proposed method, we conduct experiments on three widely recognized natural scene benchmark datasets: CAVE, NTIRE2022 , and HARVARD .

- **NTIRE2022 Dataset:** As the most extensive dataset for the SSR task to date, it provides 1,000 paired RGB and HSI samples. Each HSI is of size $482 \times 512 \times 31$, covering the spectral range of 400-700 nm at 10 nm intervals.
- **CAVE Dataset:** This dataset consists of 31 indoor scenes captured by an Apogee Alta U260 camera, featuring diverse objects such as faces, fruits, and fabrics. Each hyperspectral image (HSI) has a spatial resolution of $512 \times 512$ and contains 31 spectral bands ranging from 400 nm to 700 nm with a 10 nm increment.
- **HARVARD Dataset:** This collection comprises 50 HSIs of both indoor and outdoor environments. The images possess a higher spatial resolution of $1392 \times 1040$, with 31 bands sampled from 420 nm to 720 nm at 10 nm steps.

Consistent with common practices, the RGB counterparts for HARVARD is synthesized using the spectral response function of a Nikon D700 camera, while the NTIRE2022 dataset provides its own original paired RGB-HSI data.

*2) Implementation Settings:* **Datasets and Partitioning.** For NTIRE2022, we follow the commonly used setting in MST++ [10] and use 950 publicly available paired RGB-HSI samples, including 900 images for training and 50 images for validation. For CAVE, 24 scenes are randomly selected for training and the remaining 7 scenes are used for testing. For Harvard, we follow an 8:2 split, using 40 images for training and 10 images for validation. Since Harvard does not provide paired RGB observations, its RGB inputs are synthesized from the ground-truth HSIs using the Nikon D700 spectral response function while maintaining 16-bit precision.

**Data Augmentation.** During training, each paired RGB-HSI sample is randomly cropped into $128 \times 128$ patches and augmented by random rotations of 90°, 180°, and 270°, as well as horizontal and vertical flipping.

**Implementation Details.** Our proposed R2H-Diff is implemented using the PyTorch framework on an NVIDIA RTX 3090 GPU. We employ the AdamW optimizer with a mini-batch size of 20, except for Restormer, which uses a mini-batch size of 8 due to GPU memory limitations. The initial learning rate is set to $4 \times 10^{-4}$ and is dynamically adjusted via the CosineAnnealingLR scheduler. All training samples are cropped into $128 \times 128$ patches, while both training and validation are performed on the central $256 \times 256$ region of each image. Following standard practice in diffusion-based reconstruction, our model is configured to predict the clean HSI $x_0$ directly to ensure stable spectral fidelity.

*3) Evaluation Metrics:* Following common practice in spectral super-resolution [10], [16], [18], [41], we adopt four standard quantitative metrics: Mean Relative Absolute Error (MRAE), Root Mean Square Error (RMSE), Peak Signal-to-Noise Ratio (PSNR), and Spectral Angle Mapper (SAM). MRAE and RMSE evaluate reconstruction errors, PSNR measures spatial fidelity, and SAM reflects spectral angular consistency. For MRAE, RMSE, and SAM, lower values indicate better performance, whereas higher PSNR denotes better reconstruction quality.

**Algorithm 1** Training the Proposed Model

1: **Input:** HSI ground truth $\mathbf{x}_0$, conditional RGB image $\tilde{\mathbf{x}}$.
2: **Output:** Optimized $\mathbf{x}_0$-prediction model $f_\theta$.
3: **repeat**
4: $(\tilde{\mathbf{x}}, \mathbf{x}_0) \sim q(\tilde{\mathbf{x}}, \mathbf{x})$
5: $t \sim \text{Uniform}(\{1, \dots, T\})$
6: $\boldsymbol{\epsilon} \sim \mathcal{N}(\mathbf{0}, \mathbf{I})$
7: $\mathbf{x}_t = \sqrt{\bar{\alpha}_t}\mathbf{x}_0 + \sqrt{1 - \bar{\alpha}_t}\boldsymbol{\epsilon}$
8: $\hat{\mathbf{x}}_\theta = f_\theta(\mathbf{x}_t, t, \tilde{\mathbf{x}})$
9: Take a gradient descent step on:
10: $\nabla_\theta \left( \|\mathbf{x}_0 - \hat{\mathbf{x}}_\theta\|_2^2 + \lambda \sum_{d \in \{h,v\}} \|\nabla_d \mathbf{x}_0 - \nabla_d \hat{\mathbf{x}}_\theta\|_2^2 \right)$
11: **until** converged

**Algorithm 2** Inference via Iterative Refinement

1: **Input:** Conditional RGB image $\tilde{\mathbf{x}}$, Gaussian noise $\mathbf{x}_T$.
2: **Output:** Reconstructed HSI $\mathbf{x}_0$.
3: $\mathbf{x}_T \sim \mathcal{N}(\mathbf{0}, \mathbf{I})$
4: **for** $t = T, \dots, 1$ **do**
5: $\hat{\mathbf{x}}_\theta = f_\theta(\mathbf{x}_t, t, \tilde{\mathbf{x}})$
6: **if** $t > 1$ **then**
7: $\boldsymbol{\epsilon}_\theta = (\mathbf{x}_t - \sqrt{\bar{\alpha}_t}\hat{\mathbf{x}}_\theta)/\sqrt{1 - \bar{\alpha}_t}$
8: $\mathbf{x}_{t-1} = \sqrt{\bar{\alpha}_{t-1}}\hat{\mathbf{x}}_\theta + \sqrt{1 - \bar{\alpha}_{t-1}}\boldsymbol{\epsilon}_\theta$
9: **else**
10: $\mathbf{x}_0 = \hat{\mathbf{x}}_\theta$
11: **end if**
12: **end for**
13: **return** $\mathbf{x}_0$

TABLE I: QUANTITATIVE COMPARISON WITH STATE-OF-THE-ART METHODS ON NTIRE2022 DATASET. Bold and underlined indicate the best and the second best results, respectively. The percentage in blue represents the relative improvement of our R2H-Diff over the competitive MST++.

| | Complexity | | Reconstruction Quality | | | |
|---|---|---|---|---|---|---|
| Method | Params (M) ↓ | FLOPS (G) ↓ | MRAE ↓ | RMSE ↓ | PSNR ↑ | SAM ↓ |
| HSCNN+ | 4.65 | 304.45 | 0.3814 | 0.0588 | 26.36 | 6.2228 |
| EDSR | 2.42 | 158.32 | 0.3277 | 0.0437 | 28.29 | 5.6321 |
| HDNet | 2.66 | 173.81 | 0.2048 | 0.0317 | 32.13 | 5.7640 |
| MIRNet | 3.75 | 42.95 | 0.1890 | 0.0274 | 33.29 | 5.4080 |
| Restormer | 15.11 | 93.77 | 0.1833 | 0.0274 | 33.40 | 5.6281 |
| MST++ | <u>1.62</u> | <u>23.05</u> | **0.1645** | <u>0.0248</u> | <u>34.32</u> | **4.8793** |
| **R2H-Diff (Ours)** | **0.58** (↓64%) | **12.25** (↓47%) | <u>0.1687</u> | **0.0224** (↓10%) | **35.37** (↑1.05dB) | <u>5.3639</u> |

### B. Results and Analysis

To evaluate R2H-Diff, we compare it with representative state-of-the-art methods, including HSCNN+ [16], EDSR [17], HDNet [18], MIRNet [19], Restormer [20], and MST++ [10]. These baselines span CNN-based, residual, restoration-oriented, and Transformer-based reconstruction paradigms. All methods are retrained under the same experimental protocol, and the best validation checkpoint of each method is used for final testing.

*1) Quantitative Results:* The quantitative comparison results on the NTIRE2022, CAVE, and Harvard datasets are reported in Tables I, II, and III, respectively. Overall, R2H-Diff achieves competitive or superior reconstruction performance across all benchmark datasets, demonstrating strong reconstruction accuracy and favorable generalization capability.

On NTIRE2022, as shown in Table I, R2H-Diff achieves the best RMSE and PSNR while maintaining the lowest model complexity. Compared with MST++, R2H-Diff reduces the parameter count from 1.62M to 0.58M and the FLOPs from 23.05G to 12.25G, corresponding to 64% and 47% reductions, respectively. Meanwhile, it improves RMSE from 0.0248 to 0.0224 and PSNR from 34.32 dB to 35.37 dB, showing a favorable accuracy–efficiency trade-off.

On the CAVE dataset, Table II shows that R2H-Diff obtains the best MRAE and RMSE, reaching 0.1430 and 0.0118, respectively. Although MST++ achieves the highest PSNR and HDNet obtains the best SAM, R2H-Diff still delivers competitive performance on these metrics, indicating its ability to preserve both spectral fidelity and image-level reconstruction quality.

For the Harvard dataset, as reported in Table III, R2H-Diff achieves the best results across all four metrics. It obtains 0.1333 MRAE, 0.0099 RMSE, 40.9151 dB PSNR, and 6.9135 SAM, clearly outperforming the compared methods. These results demonstrate the robustness of R2H-Diff under more complex real-world illumination and material variations.

*2) Qualitative Comparison:* To further evaluate the reconstruction quality of the proposed method, we provide qualitative comparisons using band-wise difference maps and spectral response curves. The difference maps visualize the absolute reconstruction error between the reconstructed hyperspectral image and the ground truth at representative wavelengths. In these maps, darker blue regions indicate lower reconstruction errors, while warmer colors correspond to larger deviations.

Figure 6 compares band-wise error maps on NTIRE2022. R2H-Diff produces fewer high-error regions than competing

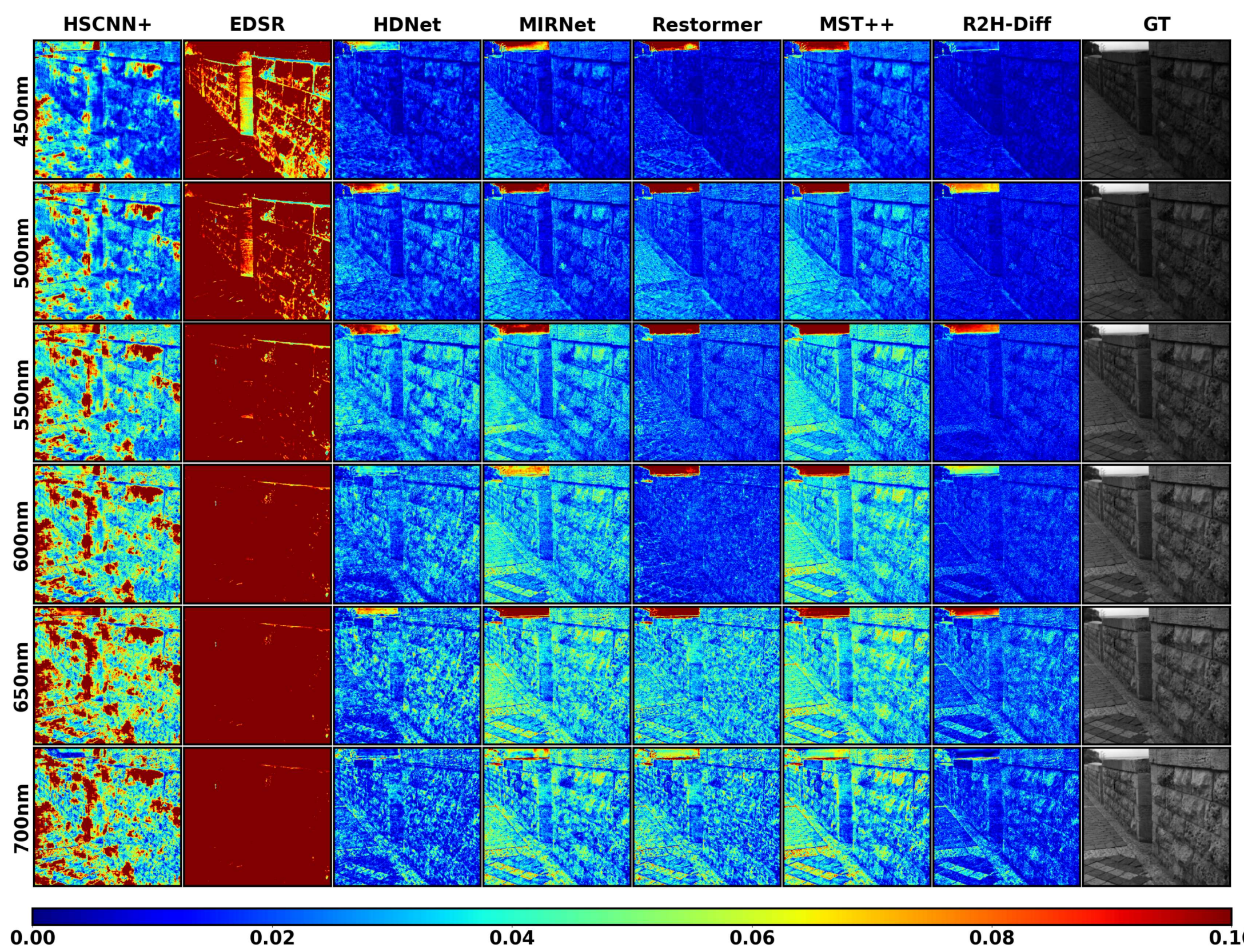

Fig. 6: Difference map comparison of scene ARAD_1K_0909 from the NTIRE2022 dataset at six representative spectral bands.

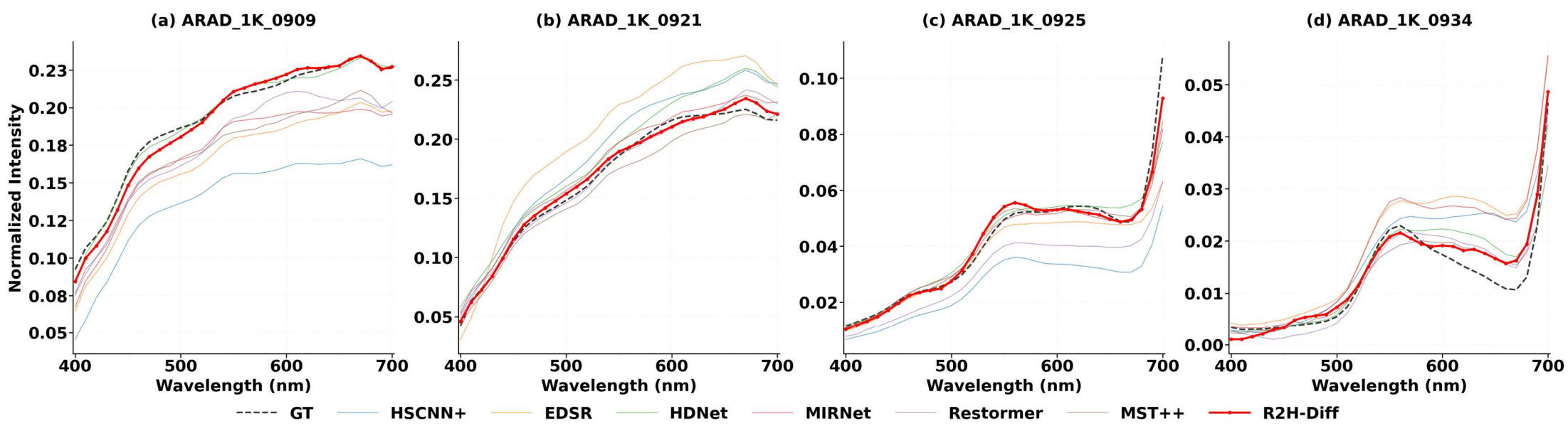

Fig. 7: Spectral response curves at randomly selected points from representative NTIRE2022 and CAVE scenes. Solid and dashed lines denote the spectra reconstructed by R2H-Diff and the ground truth, respectively.

TABLE II: Quantitative comparison with state-of-the-art methods on the CAVE dataset. Bold and underlined indicate the best and the second-best results, respectively.

| Method | MRAE ↓ | RMSE ↓ | PSNR ↑ | SAM ↓ |
|---|---|---|---|---|
| HSCNN+ | 0.8564 | 0.0434 | 27.5683 | 13.1466 |
| EDSR | 0.3384 | 0.0178 | 35.7096 | 11.7780 |
| HDNet | <u>0.1803</u> | 0.0135 | 37.9533 | **8.6934** |
| MIRNet | 0.1990 | 0.0142 | 37.6209 | 9.6321 |
| Restormer | 0.2745 | 0.0152 | 36.9124 | 11.6049 |
| MST++ | 0.2276 | <u>0.0119</u> | **39.1215** | <u>8.9583</u> |
| **R2H-Diff** | **0.1430** | **0.0118** | <u>38.9614</u> | 9.5758 |

TABLE III: Quantitative comparison with state-of-the-art methods on the Harvard dataset. Bold and underlined indicate the best and the second-best results, respectively.

| Method | MRAE ↓ | RMSE ↓ | PSNR ↑ | SAM ↓ |
|---|---|---|---|---|
| HSCNN+ | 0.4225 | 0.0233 | 33.1759 | <u>7.6366</u> |
| EDSR | 0.3739 | 0.0205 | 34.1853 | 21.1596 |
| HDNet | 0.2603 | 0.0148 | 37.0575 | 15.4538 |
| MIRNet | 0.2132 | 0.0143 | 38.2607 | 9.3711 |
| Restormer | 0.5302 | 0.0218 | 33.4337 | 19.0397 |
| MST++ | <u>0.1939</u> | <u>0.0136</u> | <u>38.6992</u> | 8.2222 |
| **R2H-Diff** | **0.1333** | **0.0099** | **40.9151** | **6.9135** |

methods, especially in textured and boundary regions, indicating better preservation of spatial-spectral structures. Figure 8

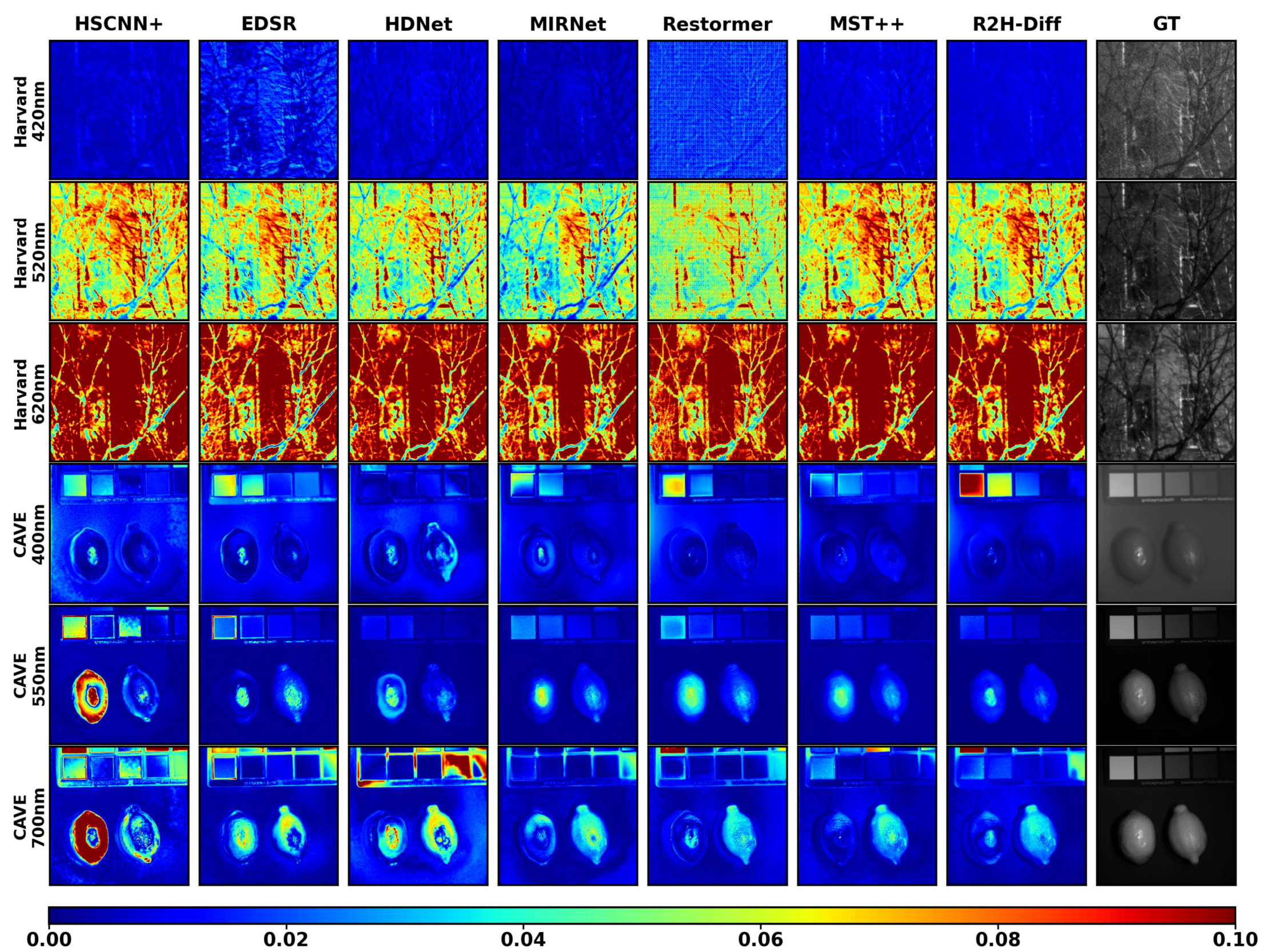

Fig. 8: Difference map comparison on the Harvard and CAVE datasets. The first three rows present the reconstruction errors at 420 nm, 520 nm, and 620 nm on Harvard, while the last three rows show the results at 400 nm, 550 nm, and 700 nm on CAVE. Warmer colors indicate larger reconstruction errors. The last column shows the corresponding ground-truth spectral bands.

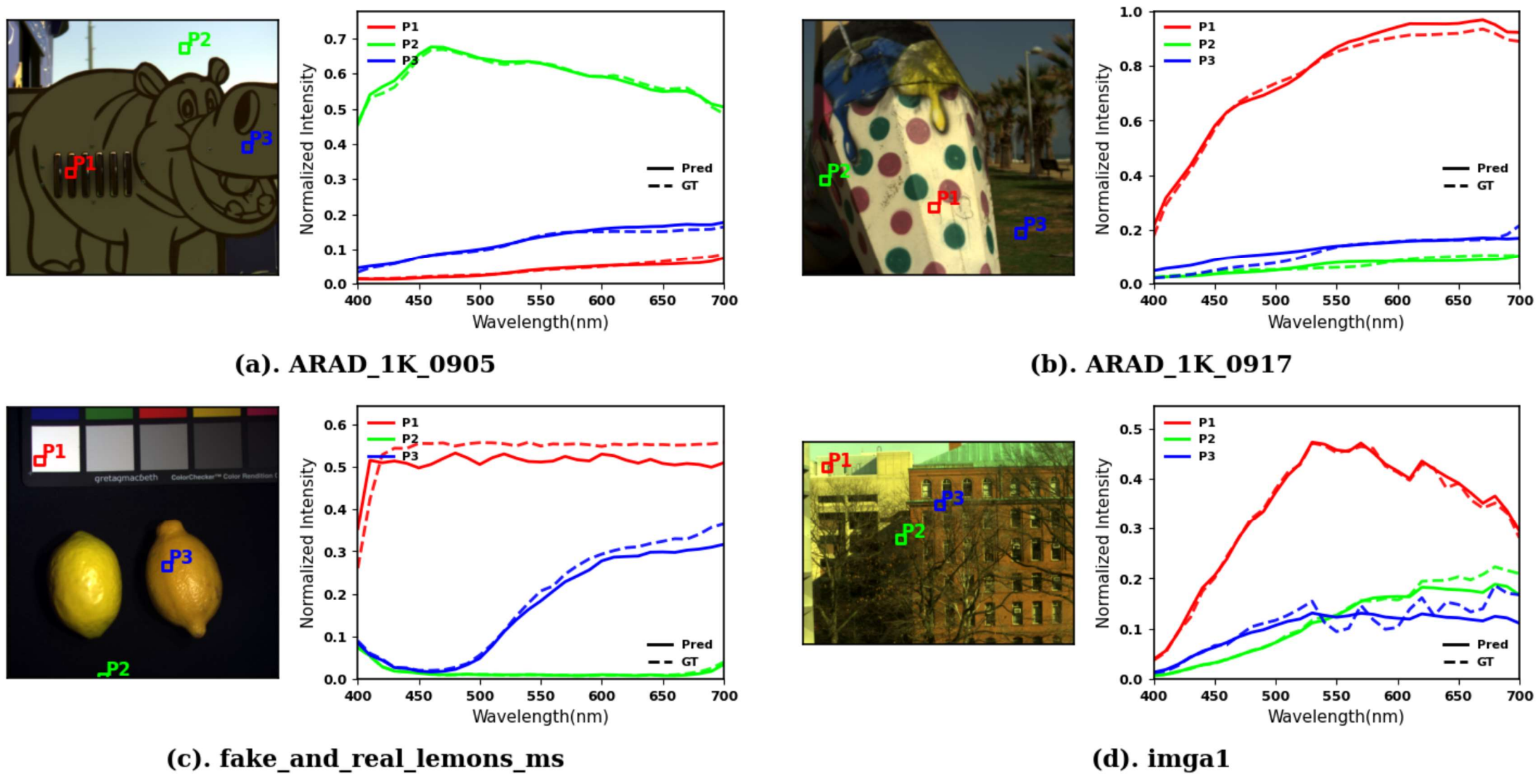

Fig. 9: Visual comparison of spectral response curves on representative scenes from multiple hyperspectral image reconstruction benchmarks. Scenes (a) and (b) are selected from the NTIRE 2022 dataset, scene (c) is selected from the CAVE dataset, and scene (d) is selected from the Harvard dataset.

further reports results on Harvard and CAVE scenes. Under complex illumination, shadows, and material-dependent variations, R2H-Diff generally yields cleaner and more stable error maps across wavelengths. The spectral curves in Fig. 9 also show closer alignment with the ground truth, demonstrating improved spectral fidelity. Overall, the qualitative results are consistent with the quantitative comparisons and verify the robustness of R2H-Diff across diverse datasets.

### C. Ablation Studies

We conduct ablation studies on the NTIRE2022 validation set to verify the effects of GSRM, HATA, and the gradient consistency loss. All variants are trained from scratch under the same experimental protocol. The full R2H-Diff model is used as the reference setting.

*1) Effectiveness of Key Components:* Table IV reports the results of removing GSRM and HATA from the proposed framework. When GSRM is removed, the model suffers consistent degradation on all metrics, with MRAE increasing from 0.1687 to 0.1842 and PSNR decreasing from 35.37 dB to 34.81 dB. This indicates that simply feeding RGB information into the denoising network is insufficient, while the proposed GSRM provides more effective RGB-conditioned spectral guidance for the reconstruction process.

The degradation becomes significant when HATA is removed. Specifically, MRAE and RMSE increase to 0.4156 and 0.0443, respectively, while PSNR drops sharply to 29.18 dB. This result demonstrates that spatial–spectral dependency modeling is critical for RGB-to-HSI reconstruction, since hyperspectral bands exhibit strong inter-band correlations that cannot be captured by ordinary convolutional blocks alone. By introducing channel-wise transposed attention with local spatial refinement, HATA effectively improves spectral feature representation while maintaining computational efficiency.

Overall, the full R2H-Diff model achieves the best performance across all evaluation metrics, confirming that GSRM and HATA are complementary. GSRM enhances RGB-guided conditional fusion, whereas HATA strengthens spatial–spectral dependency modeling. Their combination enables more accurate and stable hyperspectral reconstruction.

TABLE IV: Ablation study of key architectural components on the NTIRE2022 validation set.

| Variant | MRAE ↓ | RMSE ↓ | PSNR ↑ | SAM ↓ |
|---|---|---|---|---|
| w/o GSRM | 0.1842 | 0.0232 | 34.81 | 5.4633 |
| w/o HATA | 0.4156 | 0.0443 | 29.18 | 9.2996 |
| **R2H-Diff** | **0.1687** | **0.0224** | **35.37** | **5.3639** |

*2) Effect of Gradient Consistency Loss:* Table V analyzes the influence of the gradient consistency loss weight $\lambda$. When $\lambda = 0$, the model is trained only with the reconstruction loss, resulting in inferior performance with 0.2022 MRAE, 0.0253 RMSE, 34.03 dB PSNR, and 6.1362 SAM. This suggests that pixel-wise supervision alone cannot preserve fine spatial details, particularly near edges and textured regions.

Introducing gradient consistency supervision improves the reconstruction quality. As $\lambda$ increases from 0 to 1.0, MRAE decreases from 0.2022 to 0.1687, RMSE decreases from 0.0253 to 0.0224, and PSNR improves from 34.03 dB to 35.37 dB. These improvements indicate that gradient-domain constraints help preserve structural details and alleviate over-smoothed reconstructions. Although $\lambda = 0.5$ achieves the best SAM, its PSNR and RMSE are inferior to those obtained with $\lambda = 1.0$.

When $\lambda$ is further increased to 2.0, the performance drops on all metrics compared with $\lambda = 1.0$. This shows that an overly large gradient weight may overemphasize structural consistency and disturb spectral intensity reconstruction. Therefore, $\lambda = 1.0$ provides the best overall trade-off between spectral accuracy and spatial structure preservation, and is adopted in all experiments.

TABLE V: Effect of the gradient consistency loss weight $\lambda$ on the NTIRE2022 validation set.

| $\lambda$ | MRAE ↓ | RMSE ↓ | PSNR ↑ | SAM ↓ |
|---|---|---|---|---|
| 0 | 0.2022 | 0.0253 | 34.03 | 6.1362 |
| 0.1 | 0.1987 | 0.0267 | 33.65 | 6.0246 |
| 0.5 | 0.1696 | 0.0234 | 34.62 | **5.1519** |
| 1.0 | **0.1687** | **0.0224** | **35.37** | 5.3639 |
| 2.0 | 0.1816 | 0.0244 | 34.45 | 5.4876 |

## V. Conclusion

In this paper, we proposed R2H-Diff, an efficient diffusion-based framework for RGB-to-HSI reconstruction. By formulating spectral recovery as a conditional iterative refinement process, R2H-Diff better handles the ill-posed RGB-to-spectral mapping while maintaining high reconstruction efficiency. The proposed GSRM, HATA module, normalization-free backbone, and gradient consistency loss jointly improve RGB-conditioned spectral fusion, spatial-spectral dependency modeling, and structural preservation. Extensive experiments on NTIRE2022, CAVE, and Harvard demonstrate that R2H-Diff achieves competitive or superior reconstruction accuracy with lightweight model complexity. Future work will explore sensor-aware diffusion strategies to further improve generalization under different camera response functions.

## Acknowledgments

This work was supported by the Project of Laboratory of Advanced Agricultural Sciences, Heilongjiang Province (Grant No. ZY04JD05-010), the National Natural Science Foundation of China (Grant No. 62350710797 and No. 62103161), the Key Research and Development Program of Heilongjiang Province (Grant No. JD2023GJ01-01), and the Key Research and Development Program of Heilongjiang Province (Grant No. 2023ZX01A24).